\documentclass[a4paper]{article}

\usepackage{INTERSPEECH2021}

\usepackage{graphicx}
\usepackage[utf8]{inputenc}
\usepackage{multirow}

\usepackage{soulutf8}
\usepackage{todonotes}
\usepackage{placeins}
\usepackage{tikz}
\usepackage{svg}
\usepackage{hyperref}

\newcommand\citep{\cite}

\title{CORAA: a large corpus of spontaneous and prepared speech manually validated for speech recognition in Brazilian Portuguese}

\name{Arnaldo Candido Junior$^1$, Edresson Casanova$^2$, Anderson Soares$^3$, Frederico Santos de Oliveira$^3$, 
Lucas Oliveira$^1$, 
Ricardo Corso Fernandes Junior$^1$,
Daniel Peixoto Pinto da Silva$^1$, 
Fernando Gorgulho Fayet$^2$,
Bruno Baldissera Carlotto$^2$,
Lucas Rafael Stefanel Gris$^1$,
Sandra Maria Aluísio$^2$
}

\address{
    $^1$ Federal University of Technology – Paraná, Brazil \\
    $^2$ Instituto de Ciências Matemáticas e de Computação, University of S\~ao Paulo, Brazil \\
    $^3$ Federal University of Goias, Brazil
    }
 \email{arnaldocan at gmail dot com }%

\begin{document}

\maketitle
\begin{abstract}
 Automatic Speech recognition (ASR) is a complex and challenging task. In recent years, there have been significant advances in the area.
In particular, for the Brazilian Portuguese (BP) language, there were about 376 hours public available for ASR task until the second half of 2020. With the release of new datasets in early 2021, this number increased to 574 hours. 
The existing resources, however, are composed of audios containing only read and prepared speech. There is a lack of datasets including spontaneous speech, which are essential in different ASR applications.
This paper presents CORAA (Corpus of Annotated Audios) v1. with 290.77 hours, a publicly available dataset for ASR in BP containing validated pairs (audio-transcription). CORAA also contains European Portuguese audios (4.69 hours). We also present a public ASR model based on Wav2Vec 2.0 XLSR-53 and fine-tuned over CORAA. Our model achieved a Word Error Rate of 24.18\% on CORAA test set and 20.08\% on Common Voice test set. When measuring the Character Error Rate, we obtained 11.02\% and 6.34\% for CORAA and Common Voice, respectively.
CORAA corpora were assembled to both improve ASR models in BP with phenomena from spontaneous speech and motivate young researchers to start their studies on ASR for Portuguese.  All the corpora are publicly available at \url{https://github.com/nilc-nlp/CORAA} under the CC BY-NC-ND 4.0 license.

\end{abstract}

\section{Introduction}
\label{intro}

Automatic Speech Recognition (ASR) is a complex and challenging. Significant progress in techniques, models for the task had occurred in recent years. The main reasons for this progress include (but are not limited to) the availability of large-scale datasets and advances in deep learning methods running over powerful computing platforms.


Despite significant advances in ASR benchmarking solutions, the main and large datasets available for training and evaluating ASR systems are English due to the predominance of the language in science and business, although there are some current efforts to build multilingual speech corpora \citep{ardila-etal-2020-common, wang-etal-2020-covost, wang2020covost, Pratap_2020}. Another problem is the environment of the recording, mostly composed of clean speech.  Regarding the style of speaking, they are read speech, such as \citep{Panayotov2015LibrispeechAA, Pratap_2020, ardila-etal-2020-common, wang-etal-2020-covost, zanonboito-EtAl:2020:LREC} or prepared speech like \citep{HernandezNGTE18, salesky2021mTeDx}. 

%


In this paper, we focus on a specific language -- the Brazilian Portuguese (BP) --, which was struggling with only a few dozen hours of public data available until the middle of 2020. The previous open dataset to train speech models in BP were much smaller than American English datasets, with only 10 hours for speech synthesis (TTS)\footnote{\url{https://github.com/Edresson/TTS-Portuguese-Corpus}} and 60 hours for ASR. The resource commonly used to train ASR models for BP is an ensemble of four small, non-conversational speech datasets: the Common Voice Corpus version 5.1 (Mozilla)\footnote{\url{https://commonvoice.mozilla.org/pt/datasets}},  Sid dataset\footnote{\url{https://doi.org/10.17771/PUCRio.acad.8372}}, VoxForge\footnote{\url{http://www.voxforge.org/pt/downloads}}, and LapsBM1.4\footnote{\url{https://laps.ufpa.brfalabrasil/}}.


In the second half of 2020, three new datasets were made available: (i) the BRSD v2 which includes  the CETUC dataset \citep{Alencar_2008} (with almost 145 hours),  plus 12 hours and 30 minutes  of  non-conversational  speech from 3 small open  datasets\footnote{BRSD v2 also includes CSLU: Spoltech Brazilian Portuguese Version 1.0 --- \url{https://catalog.ldc.upenn.edu/LDC2006S16}} \citep{Macedo_Quintanilha_2020}, (ii) the Multilingual LibriSpeech (MLS), derived from reading LibriVox audiobooks in 8 languages, including BP \citep{Pratap_2020} with 169 hours, and (iii) the dataset Common Voice version 6.1\footnote{\url{https://commonvoice.mozilla.org/pt/datasets}}  \citep{ardila-etal-2020-common}, with 50 validated hours, composed of recordings of read sentences which were displayed on the screen.  These three datasets total 376 hours. Given this recent public availability of large audio databases for BP language, the lack of resources has been gradually reduced, although it is still far from ideal when compared to resources for the English language.


In early 2021, a new dataset with prepared speech, called the Multilingual TEDx Corpus \citep{salesky2021mTeDx}, was made publicly available, providing 765 hours  to support speech recognition and speech translation research. The Multilingual TEDx Corpus is composed by a collection of audio recordings from TEDx talks in 8 source languages, including 164 hours of Portuguese. Moreover, a new version of the dataset Common Voice (Common Voice Corpus 7.0) was launched with 84 validated hours, which is an increment of 34 hours over the previous version. Therefore, currently, BP language is well represented with 574 hours of speech data which can be used to train new ASR models. 


However, there is still a lack of datasets with audio files that record spontaneous speech of various genres, from interviews to informal dialogues and conversations, i.e., conversational speech recorded in natural contexts and noisy environments to train robust ASR systems. Spontaneous speech presents several phenomena such as laughter, coughs, filled pauses, word fragments resulted from repetitions, restarts and revisions of the discourse. This gap makes difficult the development of both high-quality dialog systems and automatic speech recognition systems capable of handling spontaneous speech recorded in noisy environments. The latter ones are called rich transcription-style ASR (RT-ASR) when they explicitly convert those phenomena cited above into special tokens \citep{
inaguma17_interspeech, Fujimura2018SimultaneousSR, tanaka21c_interspeech}.  
Dialog systems, for example, must deal with  several types of speech disfluencies, preserving them instead of removing filled pauses and word fragments  \citep{BaumannKHS16}. In general, it is expected that ASR systems trained on read  style and clean speech will face a drop of performance when dealing with informal conversations in contexts of free interactions and noisy environments.

The TaRSila project is an effort of the Center for Artificial Intelligence\footnote{\url{http://c4ai.inova.usp.br/pt/nlp2-pt/}} (C4AI) to make available language resources to bring natural language processing of BP to the state-of-the-art. The project aims at growing speech datasets for BP language, to achieve state-of-the-art results for automatic speech recognition, multi-speaker synthesis, speaker identification, and  voice cloning. In a joint effort of two research centers, the C4AI and the CEIA\footnote{\url{http://centrodeia.org/}} (Center of Excellence in Artificial Intelligence, in English), four speech corpora  composed of \textbf{prepared, guided interviews and spontaneous speech} from academic projects  were manually validated to serve as an ASR benchmark for BP. The projects are:  (i) ALIP \citep{Goncalves_2019}; (ii) C-ORAL Brasil I \citep{raso_mello_2012}; (iii) Nurc-Recife \citep{Oliveira_2016}; and (iv) SP2010 \citep{MENDES_OUSHIRO_2012}. We also validated 76.36 hours of prepared speech from a collection of TEDx Talks\footnote{\url{https://www.ted.com/}} in Brazilian Portuguese, including 4.69 hours of European Portuguese, to allow experiments with Portuguese language variants.



\subsection{Goals}

In this paper we present a new public available dataset called CORAA (Corpus of Annotated Audios) v1. CORAA has 290.77 hours of validated pairs (audio-transcription) and is composed by five corpora: ALIP \citep{Goncalves_2019}, C-ORAL Brasil I \citep{raso_mello_2012}, NURC-Recife \citep{Oliveira_2016}, SP2010 \citep{MENDES_OUSHIRO_2012}, TEDx Portuguese talks. Information about each corpora is presented in Table \ref{tab:original}. The original sizes of each dataset in hours are presented as reported in their respective original papers, when reported by the authors. Regarding SP2010, the total duration is estimated, since the authors report 60 recordings from 60 to 70 minutes each and the total hours of ALIP was computed after download. All the corpora are publicly available\footnote{Currently, only the test set is not available, because it will be released after an ASR Challenge involving CORAA} at \url{https://github.com/nilc-nlp/CORAA} under the CC BY-NC-ND 4.0 license. These corpora were assembled with the purpose of improving ASR models in BP with phenomena from spontaneous speech and noise in order to motivate young researchers in this exciting research area.

\begin{table*}[!ht]
\centering
\caption{Speech Genres, Accents, Speaking Styles and Hours (in decimal) in each original CORAA Corpora}
\begin{tabular}{llllll}
\hline
\multicolumn{1}{l|}{\textbf{Corpus}} &
  \multicolumn{1}{l|}{\textbf{ALIP}} &
  \multicolumn{1}{l|}{\textbf{\begin{tabular}[c]{@{}l@{}}C-ORAL\\ Brasil I\end{tabular}}} &
  \multicolumn{1}{l|}{\textbf{\begin{tabular}[c]{@{}l@{}}NURC\\ Recife\end{tabular}}} &
  \multicolumn{1}{l|}{\textbf{SP2010}} &
  \multicolumn{1}{l}{\textbf{\begin{tabular}[c]{@{}l@{}}TEDx\\ Portuguese\end{tabular}}} \\ \hline
\multicolumn{1}{l|}{\begin{tabular}[c]{@{}l@{}}Speech \\ Genres\end{tabular}} &
  \multicolumn{1}{l|}{\begin{tabular}[c]{@{}l@{}}Interviews,\\ Dialogues\end{tabular}} &
  \multicolumn{1}{l|}{\begin{tabular}[c]{@{}l@{}}Monologues,\\ Dialogues,\\ Conversations\end{tabular}} &
  \multicolumn{1}{l|}{\begin{tabular}[c]{@{}l@{}}Dialogues,\\ Interviews,\\ Conference   \\ and Class \\ Talks \end{tabular}} &
  \multicolumn{1}{l|}{\begin{tabular}[c]{@{}l@{}}Conversations,\\ Interviews,\\ Reading\end{tabular}} &
  \multicolumn{1}{l}{\begin{tabular}[c]{@{}l@{}}Stage\\ Talks\end{tabular}} \\ \hline
  \multicolumn{1}{l|}{\begin{tabular}[c]{@{}l@{}}Speaking \\ Styles\end{tabular}} &
  \multicolumn{1}{l|}{\begin{tabular}[c]{@{}l@{}}Spontaneous \\ Speech\end{tabular}} &
  \multicolumn{1}{l|}{\begin{tabular}[c]{@{}l@{}}Spont. \\ Speech\end{tabular}} &
  \multicolumn{1}{l|}{\begin{tabular}[c]{@{}l@{}}Spont. and  \\   Prepared \\ Speech\end{tabular}} &
  \multicolumn{1}{l|}{\begin{tabular}[c]{@{}l@{}}Spont. and  \\  Read Speech\end{tabular}} &
  \multicolumn{1}{l}{\begin{tabular}[c]{@{}l@{}}Prepared\\ Speech\end{tabular}} \\ \hline
\multicolumn{1}{l|}{Accent} &
  \multicolumn{1}{l|}{\begin{tabular}[c]{@{}l@{}}São Paulo \\ State Cities\end{tabular}} &
  \multicolumn{1}{l|}{Minas Gerais} &
  \multicolumn{1}{l|}{Recife} &
  \multicolumn{1}{l|}{\begin{tabular}[c]{@{}l@{}}São Paulo \\ Capital\end{tabular}} &
  \multicolumn{1}{l}{Misc.} \\ \hline
\multicolumn{1}{l|}{\begin{tabular}[c]{@{}l@{}}Original \\ (hrs)\end{tabular}} &
  \multicolumn{1}{l|}{78} &
  \multicolumn{1}{l|}{21.13} &
  \multicolumn{1}{l|}{279} &
  \multicolumn{1}{l|}{65} &
  \multicolumn{1}{l}{249} \\ \hline
 & & & &
\end{tabular}
\label{tab:original}
\end{table*}

{As an example of the feasibility of speech recognition with CORAA, we present one speech recognition experiment using the Wav2vec 2.0 XLSR-53 \citep{conneau2020unsupervised, baevski2020wav2vec}. Furthermore, we compared our model with the state of the art in automatic speech recognition in Brazilian Portuguese\citep{gris2021brazilian}. This two models are evaluated according to three main scenarios: (a) testing audios with different characteristics from training; (b) focusing on model performance for each of the five corpora, considering noise level and accent; (c) analyzing spontaneous and prepared speech styles impacts on the trained models.}


\subsection{Highligths}

The main contributions made in this work are summarised as follows.
\begin{enumerate}
\item A large BP corpus of validated pairs (audio-transcription)
containing 290.77 hours, composed of five corpora (ALIP, C-ORAL Brasil I, NURC Recife, SP2010, and TEDx Portuguese talks), adapted for the task of ASR in BP. We also include 4.69 hours of European Portuguese (in TEDx Portuguese).
\item The first corpus, according to our knowledge, tackling spontaneous speech for ASR in BP.
\item {An ASR Model, publicly available, using the presented corpus.}
\end{enumerate}

Section 2 details both related work on datasets available for ASR in BP and the five spoken corpora projects used in CORAA v1. Section 3 describes the steps followed in preparing the CORAA corpus. {Section 4 presents the creation of train, development and test splits of CORAA, the experiment on ASR for BP and an error analysis of our model. Finally, Section \ref{sec:conc} presents the final remarks of the work.}

\section{Related Works on Speech Datasets and Spoken Corpora for BP}

\subsection{Open Datasets for Speech Recognition in BP}
\label{sec:open_bp}


Three new datasets were released for BP at the end of 2020. CETUC dataset \citep{Alencar_2008} contains 145 hours of 100 speakers, half males, and half females. The sentence set is composed of 1,000 sentences (3,528 words). The sentences are phonetically balanced and extracted from CETEN-Folha\footnote{\url{https://www.linguateca.pt/cetenfolha/index\_info.html}} corpus. Each speaker uttered all sentences from the sentence set exactly once. CETUC was recorded in a controlled environment, using a sample rate of 16kHz. The audios are publicly available\footnote{\url{https://igormq.github.io/datasets/}}, without an explicit license.  Regarding the environment of recording and speaking style, CETUC delivers clean and read speech.

Common Voice Corpus 6.1, version pt\_63h\_2020-12-11, contains 63 hours of audio, 50 of which were considered validated. The dataset comprises 1,120 BP speakers, 81\% males and 3\% females (some audios are not sex labeled). The audios were collected using the Common Voice website\footnote{\url{https://commonvoice.mozilla.org/pt/speak}} or using a mobile APP. The speakers read aloud sentences presented on the screen. A maximum of 3 contributors analyzed each pair audio-transcription, and simple voting is applied: two votes for acceptance validate the audio; two votes for rejection invalidate the audio. A given release may also contain samples that were analyzed but did not receive enough votes to be validated/invalidated --- these samples have the status ``OTHER'' \citep{ardila-etal-2020-common}.
Releases are distributed under the CC-0\footnote{\url{https://commonvoice.mozilla.org/pt/datasets}} license and contain MP3 files, originally collected at 48kHz sampling rate but downsampled to 16kHz. The following metadata is also available: ID\_speaker, path\_audio\_file, read\_sentence, up\_votes, down\_votes, age, gender, accent. Where up\_votes e down\_votes refers to the voting result, and the last three fields are optional.
Regarding the speaking style, Common Voice Corpus has read speech. As for recording environment, both noise level and sound clarity is very heterogeneous. The current version of the dataset  (Common Voice Corpus 7.0) has 84 validated hours, 34 hours more than version 6.1.

The Multilingual LibriSpeech (MLS) dataset \citep{Pratap_2020} is composed by audios extracted from  Librivox\footnote{\url{https://librivox.org/}} audiobooks. The Librivox project releases audiobooks in the public domain. MLS dataset encompasses eight languages, including BP, and is released under the CC BY 4.0\footnote{\url{http://www.openslr.org/94/}} license. MLS can be used for developing both ASR  and TTS models. There are 160.96 hours for training models, 3.64 hours for tuning and 3.74 for testing for Portuguese. It provides 26 male and 16 female speakers in the training dataset; 5 female, and 5 male speakers for tuning; and the same for testing. The audios were downsampled from 48kHz to 16kHz for easy processing. Regarding the environment of the recording and speaking style, MLS is made of clean and read speech.

In early 2021, a new dataset was made publicly available --- the Multilingual TEDx Corpus, licensed under the CC BY-NC-ND 4.0\footnote{www.openslr.org/100}. This dataset has recordings of TEDx talks in 8 languages, BP being one of them, represented with 164 hours and 93K sentences. Each TEDx talk is stored as a 44 or 48kHz sampled wav file. Available metadata include source language, talk title, speaker name, audio length, keywords, and a short talk description. Multilingual TEDx Corpus was built to advance ASR and speech translation research, with multilingual models and baseline models being distributed for ASR and speech translation. Regarding the speaking style and the environment of the recording, Multilingual TEDx Corpus is composed of prepared and clean speech. 

\subsection{Spoken corpora projects used in CORAA}
\label{sec:coraa-subdatasets}

\subsubsection{ALIP} 


The project ALIP\footnote{\url{https://www.alip.ibilce.unesp.br/}} (Amostra Linguística do Interior Paulista -- Language Sample of the Interior of São Paulo, in English) \citep{Goncalves_2019} was proposed in $2002$, and coordinated by Prof. Sebastião Carlos Leite Gonçalves, from UNESP São José do Rio Preto. This project was responsible for building the database called Iboruna \citep{IBORUNA}, composed of two types of speech samples:


\begin{itemize}
     \item A sample of $151$ interviews (each with about $20$ minutes, being $76$ male and $76$ female voices) 
     from the northwest region of the  São Paulo state;
     \item Another sample consisting of $11$ dialogues, involving from two to five informants. It was secretly recorded in contexts of free social interactions. This sample has 28 informants (10 men and 18 women).
\end{itemize}




This corpus totals $78$ hours and it is characterized by the spontaneous speech of the linguistic variety of Brazilian Portuguese spoken in the interior of São Paulo. It was compiled between the years of $2004$ and $2005$. The informants, residents of $7$ different cities, range in age from $7$ to over $55$ years, with a considerable variety of income and education.

The speech samples were recorded with GamaPower and PowerPack digital recorders. For interviews, the consent of the informants was obtained before recording, while, for the dialogues, dialogues, the consent was obtained after recording. The interviewer conducted the interviews, and the dialogues were free, with topics defined by the participant interactions.

The corpus is available for academic use without a defined license, but with defined Terms of Use and Privacy Policy\footnote{\url{https://www.alip.ibilce.unesp.br/termos-de-uso}}. It is available via download from the project website. The two types of samples have a dedicated folder for each, in the following formats. Each folder contains .mp3 files (the audios are sampled in 8kHz), as well as .doc and .pdf files (transcriptions, informant's socio-demographic information, among others). It is important to note that audio files are not aligned with their transcriptions. 

 \subsubsection{C-ORAL Brasil I} 


C-ORAL Brasil I is a corpus published in $2012$, resulting of the project C-ORAL Brasil\footnote{\url{http://www.c-oral-brasil.org/}} under the coordination of Tommaso Raso and Heliana Mello, from the Faculty of Arts of the Federal University of Minas Gerais \citep{Raso_Mello_Mittmann_2015,raso-etal-2012, raso_mello_2012}. This synchronic corpus was recorded between 2008 and 2011 and is composed of informal and spontaneous speech, representative of the linguistic variation in Minas Gerais, especially in the city of Belo Horizonte.

 

It is composed of 139 recording sessions (or texts), totaling $21.13$ hours and $208,130$ words, averaging 1,500 words per text. C-ORAL Brasil I has 362 informants. There is a balance regarding number of uttered words: 50.36\% words are uttered by 159 males and 49.64\% words by 203 females.

Its content is divided into private-family (about $3/4$ of the corpus) and public ($1/4$) contexts. In addition, there is a separation of interaction types by number of participants: monologues (amounting to about $1/3$ of recordings), dialogues and conversations, i.e. more than two active participants (about $2/3$ of recordings).

 

The speech flow was segmented into tonal units and terminal units according to the prosodic criterion, based on the Language Into Act Theory (L-AcT) \citep{CRESTI2018} which designates the utterance as the reference unit of speech. The boundary between tonal units results from a prosodic break with a non-conclusive value, while the boundary
between terminal units corresponds to the perception of a prosodic break with a conclusive value.


In order to obtain a great diaphasic diversity, i.e, according to the communicative context, the project brought a remarkable variety of communicative contexts, compiling scenarios such as communication between players in a football match, the preparation of a drag queen for a presentation, a conversation between a realtor and a client, among others. In addition, a considerable balance was reached regarding the demographic criterion concerning the informants' education and gender. There are 362 informants in the corpus, 138 from the city of Belo Horizonte, 89 from other cities in Minas Gerais, and the rest from other states, countries, or of unknown origin.
 
 

There was an effort to use high-quality acoustic equipment at the time. The project used PMD660 Marantz digital recorders and Sennheiser Evolution EW100 G2 wireless kits. It also used non-invasive ``clip-on'' microphones to create a more natural environment, essential for recording high diaphasic variation in spontaneous speech.

C-ORAL Brasil I is available via download from the project website in raw format, morphosyntactically annotated by the Parser Palavras \citep{bick2000palavras}, in addition to metadata. The C-Oral-Brasil I corpus is licensed under CC BY-NC-SA 4.0.  The following files are of special interest for this work:
(i) audio in .wav format, with a sampling rate of 48kHz, transcription in .rtf and .txt formats, audio-transcription alignment in XML format generated by the software WinPitch\footnote{\url{https://www.winpitch.com/}}.

 \subsubsection{NURC-Recife} 
 

The NURC-Recife corpus has its origins in the $1969$ NURC (\textit{Norma Urbana Oral Culta}) project, which documents the spoken language in five Brazilian capitals: Recife, Salvador, Rio de Janeiro, São Paulo and Porto Alegre. NURC-Recife corresponds to the part referring to the linguistic variety spoken in the city of Recife. The corpus is available on the website of the NURC Digital project\footnote{\url{https://fale.ufal.br/projeto/nurcdigital/}}, developed between $2012$-$2016$. The project NURC Digital, coordinated by Prof. Miguel Oliveira Jr. of the Federal University of Alagoas (UFAL), was responsible for processing, organizing and releasing the data of the NURC-Recife project in digital form \citep{Oliveira_2016}.


The project is comprised of $290$ hours spread over $346$ recordings (called inquiry in the project) obtained between the years of $1974$ and $1988$. In fact, this value would be the total duration in hours if all audios and their transcriptions were available on the website. An analysis of all audio-transcription pairs raised one inquiry lacking audio and transcription and 11 inquiries lacking transcriptions,
resulting 279 hours available.

 

The recordings follow NURC guidelines and are categorized as follows:

 
\begin{itemize}
      \item Formal utterances (EF), consisting of 37 recordings of lectures and talks given by one speaker;
      \item Dialogues between two informants (D2) conducted by a mediator, with 71 recordings;
      \item Dialogues between an informant and an interviewer (DID), with 238 recordings.
  \end{itemize} 
 


The informant ages range  from $25$ to over $56$ years, all of them with higher education and initially selected with equal division (originally $300$-$300$) for male and female voices.

The environment of the recordings varied, depending on the type of inquiry: specific rooms, classrooms, auditoriums or even in the informants' homes. It also has very heterogeneous noise levels and sound clarity, whether from the equipment used, the recording environment or deterioration of the physical material.


The original recordings were captured with omnidirectional dynamic microphones with table support.
The reel-to-reel tape recorders used were: AKAI 4000 DS Mk–II, SONY TC–366, and Philips N 4416, the first being the most frequent. The audios were recorded on professional reel magnetic tapes, $0.0018$mm thick, $6.35$mm wide, and $540$m long (BASF TP 18 LH). However, within the scope of the NURC Digital project, they were digitized following the recommendations of the Open Archival Information System (OAIS), in the ISO standard ($14721:2003$), with a sampling rate of $96$kHz and quantization of $24$ bits. For this digitization, were used the software Audacity, Audiofile Specter, the AKAI 4000 DS Mk–II reel-to-reel recorder, a USB Audio Interface Sound Devices USBPre 2, and the RCA Diamond Cable JX-2055.



NURC Digital is available for academic use, without a defined license, via download from the project website, which allows a search by recording year (1974 to 1988), recording topic, and type of inquiry (D2, DID, and EF). There is also information about the age range of the informants, gender, and audio quality. Within each inquiry folder there are: (i) the digitized version of the specific recording (metadata), in .pdf format; (ii) a file in textgrid format, containing the audio timestamps with the transcriptions; (iii) the audio file of the recording in .wav format (48kHz); (iv) a copy of the audio file, also in .wav format, compressed at a frequency of 44kHz; and (v) the original transcription in .pdf format.

 \subsubsection{SP2010} 
 
 
 The SP2010 project \citep{SP2010_website, MENDES_OUSHIRO_2012} was coordinated by Prof. Ronald Beline Mendes, of the Research Group in Sociolinguistics at FFLCH/USP (GESOL-USP). It started in $2009$ and ended in $2013$ to document and study the Portuguese spoken in the city of São Paulo. The project was supported by the FAPESP agency between 2011 and 2013, generating a corpus publicly available for academic research.


The corpus contains $60$ recordings of 60 to 70 minutes each, collected between $2012$ and $2013$\footnote{\url{http://projetosp2010.fflch.usp.br/corpus}}, with equal division for female and male voices. Each recording identifies an interview with an informant, comprising two parts:
\begin{itemize}
     \item an informal and spontaneous conversation, with questions about the informant's neighborhood, family, childhood, work and leisure, seeking personal involvement;
     \item the continuation of the conversation, but exploring a more argumentative speech, with questions on more objective themes about the city of São Paulo, involving problems, solutions, characterizations of the city and its inhabitants. In addition, there are three reading recordings: a list of words, a news article and  a statement. Finally, specific questions about the sociolinguistic varieties of the city are proposed.
\end{itemize}


The informants were selected to represent $12$ sociolinguistic profiles characterized by distinct combinations of the following variations: age group, (with three age groups encompassing individuals from $19$ to $89$ years), education, (with two school stages represented --- up to elementary school and with higher education), and gender, (male and female). Each sociolinguistic profile has five informants as representatives, each with a recording.  The informants' region of residence within the city was also considered, and a balance of informants was sought in this regard, considering the division of São Paulo into 3 regions: \textit{Centro Velho, Centro Expandido} and \textit{Periferia}.



For the recording, the authors used TASCAM DR100 MK2 digital recorders and Sennheiser HMD25-1 microphones, having varied recording conditions, with some interviews being more noisy than others, as they were not conducted in specialized and isolated environments.

The material collected in the SP2010 project is made available via download from the project website, free of charge to the academic community of researchers.
Eight files are available for each interview: two audio files --- in .wav stereo format, 44kHz, and also in .mp3; four transcription files (in .eaf, .doc, .txt and textGrid formats); the informant and the recording forms (in .xls format); and a .zip file that contains all of the interview materials except the .wav file.

\subsubsection{TEDx Portuguese} 
\label{TeDx}

TEDx Portuguese is a new corpus compiled specifically for CORAA v1. It should not be confounded with the BP audios available in Multilingual TEDx Corpus (described in Section \ref{sec:open_bp}). TEDx Portuguese is based on the 
TEDx Talks\footnote{\url{https://www.ted.com/watch/TeDx-talks}}, which are events in which presentations on a wide range of topics take place,  and in the same format as the TED Talks\footnote{\url{https://www.ted.com/}}, but in languages other than English. 

Although they are independent meetings, they are licensed and guided by the TED organization, that is, they are short presentations, containing prepared speech, with a duration recommendation of less than 18 minutes, typically presented by a single presenter. The ``x'' at the end indicates that the event is carried out by autonomous entities worldwide. More than 3,000 new recordings are made annually\footnote{\url{https://www.ted.com/about/programs-initiatives/TeDx-program}}.


To create this dataset, we selected presentations spoken in Portuguese, both from Brazil and Portugal, with available preexisting subtitles. After selecting the presentations, they were downloaded, were the audios were extracted and converted to .wav format, mono, with a sampling rate equal to 44kHz. BP presentations have accents from practically all regions of Brazil.


The subtitles were also downloaded, with the text extracted exclusively, that is, the timestamps were discarded. The dataset is composed of excerpts from 908 talks (671 of which are in BP), totaling at least 908 different speakers, since there are also talks with more than one speaker. 
The variant (PT-PT or PT-BR) is annotated in the dataset metadata. Considering both variants, there are 543 male and 375 female voices. 


\section{Data Processing Pipeline}


In this section, we present the processing steps of the CORAA corpus:
\begin{enumerate}
\item Normalization of transcriptions,
\item Segmentation and removal of silence and untranscribed parts of speech,
\item Forced alignment between audio and corpora transcription for two corpora\footnote{ALIP was not available in an aligned way and TEDx Portuguese were available with segmentation to optimize on-screen presentation},
\item Specific processing in the ALIP and NURC-Recife corpora. For example, (i) to maintain the capitalization of letters indicative of names, to aid in the expansion of names, (ii) to preserve the slashing annotation, indicative of truncation in the speaker's speech, to aid in the identification of truncated audios, and (iii) to discard audios with duration less than 0.3 seconds in the NURC-Recife\footnote{The original duration of the corpus (279 hours) dropped to 216 hours.},
\item Validation of audio-transcription pairs, via the web interface created in the project, so that the CORAA v1 corpus can be used for training ASR methods, and
\item Evaluation of agreement between annotators and between annotators and the gold-standard annotation, performed by a trained annotator.
\end{enumerate}


All corpora described in Section \ref{sec:coraa-subdatasets} were obtained from their respective official websites. After downloading, all transcripts were converted to .csv format and the organization of audio files was standardized. Additionally, due to the differences between the transcription rules of each corpora, text normalization was performed, described in Section \ref{sec:text-norm}. Furthermore, as the ALIP corpus does not originally have alignment between the transcription and the audio file, we performed the forced alignment between the transcription and the audio. TEDx Portuguese has the alignment provided by the subtitles, however, this alignment is limited to 42 characters per line to optimize screen display, and may not correspond to sentence boundaries, for this reason we also performed forced alignment in TEDx Portuguese. We describe the forced alignment process in these two corpora in Section \ref{sec:alinhamentoforca}. The validation of the audio-transcription pairs is presented in Section \ref{sec:braz} and the evaluation of agreement between annotators and between annotators and the gold standard annotation is presented in Section \ref{sec:kappa}. Finally, Section \ref{sec:stat} presents the statistics of the five corpora that make up CORAA,  after its pre-processing.

\subsection{Text Normalization} 
\label{sec:text-norm}



The four academic project corpora used their own transcription criteria. The oldest and most widely cited transcription standards are those of the NURC Project, which were used by NURC-Recife. NURC-Recife follows the orthographic transcription and its rules can be found in \cite{Dino_Preti_1999}. During the NURC Digital project, NURC-Recife went through new processing steps, including: quality verification of digitized audio, manual alignment between audio and transcription, spelling revision using a spell checker, which are described by 
\cite{Oliveira_2016}.

The corpus C-Oral-Brasil I follows the orthographic-based transcription criteria, but with the implementation of
some non-orthographic criteria to capture grammaticalization or lexicalization phenomena \citep{Raso_Mello_2009}. For example, there are aphereses (disappearance of a phoneme at the beginning of a word), reduced prepositions, absence of plural mark in noun phrases, cliticizations of pronouns and pre-verbal negation and articulations of preposition with article.



The SP2010 project uses semi-orthographic transcriptions, using the following criteria: (i) no change in the spelling of words, as phonetic transcription is not used; (ii) no grammatical corrections; (iii) use of parentheses to indicate the deletion of /r/ in syllabic coda, syllable /es/ of the verb ``estar'' (to be), in all tenses and verb modes, and syllable ``vo'' of ``você(s)'' (you). Other deletions were not indicated with marks. Filled pauses, interjections, and conversational markers such as ``right ?'', ``okay ?'' were pervasively used.


The ALIP project follows the orthographic conventions of the written language, but uses capital initials only for proper names. The transcription annotates the following variable phenomena \citep{Goncalves_2019}: (i) vowel raising in contexts
of medial postonic of nouns, as in ``c[o]zinha $\sim$ c[u]zinha'' and of verbs, as in ``d[e]via $\sim$ d[i]via''; (ii) postonic lifting and syncope medial, as in ``pes.s[e].go $\sim$ pes.s[i].go $\sim$ pes.go'';
(iii) gerund reduction, as in ``canta[ndo]~$\sim$~canta[no]'', a striking feature of São Paulo speech.


Results for variable phenomena of morphosyntactic order include, for example, the realization of prepositions with and without contraction, as in ``com a $\sim$ cu’a $\sim$ c’a'', ``para $\sim$ pra $\sim$ pa''. The corpus proposed a transcription system based on the NURC project and reports the transcription conventions grouped in the following criteria: (i) word spelling, which includes, for example, question and exclamation marks next to the markers discursive and interjections, use of ``/'' for word truncations; (ii) prosodic elements where it uses an ellipsis for pauses, double-typed colons for lengthening vowels, and interrogation for questions; (iii) interaction in which it identifies the participants of the interaction and use square brackets for voice overlappings; (iv) transcriber's comments where parentheses are used for hypotheses of what is heard and double parentheses for descriptive comments for laughs, for example.


Considering these differences between the transcriptions and seeking to maintain standardization, we performed the following normalizations in the texts of all CORAA corpora. Some normalizations were performed before validation (items (1), (2), (3)) and practically the entire list below was performed at the end of the entire process, since the ALIP and TEDx Portuguese corpora had their transcriptions revised:


\begin{enumerate}
    \item Removal of extra annotations that do not belong to the alignment of transcripts and audios, such as annotations that indicate the speech of the interviewer and interviewee, truncations, laughter and extra information provided by the annotators of the projects that make up CORAA corpus;
    \item Normalization of texts to lower case;
    \item Removal of duplicate spaces;
    \item Expansion of acronyms for their forms of pronunciation (standardization applied after validation, to guarantee the expansion of all acronyms);
    \item Standardization of some uses of filled pauses, using a reduced set of these: \textit{ah}, \textit{eh} and \textit{uh}. Some variations of these representations have been replaced by the closest of the three above (e.g.: \textit{hum, hm, uhm} was replaced by \textit{uh}; \textit{éh, ehm, ehn}, was replaced by \textit{eh}; \textit{huh, uh, ã} was replaced by \textit{ah});
    \item Expansion of cardinal and ordinal numbers, using the num2words library\footnote{\url{https://github.com/savoirfairelinux/num2words}};
    \item Percentage sign expansion (\%) for its transcribed form (percentage);
    \item Removal of characters such as punctuation and non-language symbols (such as parenthesis and hyphen).
\end{enumerate}


It is important to note that the corpus also brings a great variety of filled pauses forms, so that the model can learn to vary its use, although this richness penalizes the evaluation of models trained with CORAA v1 corpus, as detailed in Section \ref{sec:error}.




\subsection{Automatic Forced Alignment} 
\label{sec:alinhamentoforca}




As mentioned before, in the ALIP and TEDx Portuguese corpora the alignment between the transcripts and audio was performed using an automatic forced alignment method. For this, we use the tool Aeneas\footnote{Available at \url{http://www.readbeyond.it/aeneas}.}. This tool requires the text segmented into sentences or excerpts. 

In the ALIP corpus, the text was segmented using the annotations of pauses or hesitations, indicated by ellipses (``...'') and turn-shifts between speakers, indicated by a line break followed by the next speaker identification abbreviation, present in the original annotated corpus.

In the TEDx Portuguese corpus, the segmentation of text into sentences was performed using the punctuations present in the subtitles, if any. For this, a maximum limit of 30 words was defined for each sentence and, when this limit was reached, the sentence was divided in the point before this limit. In the case of no punctuation, the sentences were divided in an arbitrary way, for example, in silent passages, or with music, or based on variations in speech rate.

\subsection{Human Validation via Web-based Platform} 
\label{sec:braz}



The validation of audio-transcription pairs was performed in a simple web interface through two tasks: \textbf{binary annotation (VALID - INVALID)} and \textbf{transcription} to correct automatic alignment effects, as was the case with ALIP corpus, or to review manual transcripts, previously made, as was the case for the TEDx Portuguese corpus.

The \textbf{binary annotation} was carried out by: listening to an audio file that could be listened to as many times as necessary and the reading of the original transcription. The annotation was binary, that is, the pairs were classified as valid or invalid, and it was necessary to point out the reason for such choice, which provided a guide for the choice itself.



There are 3 main reasons an audio is considered invalid:
\begin{enumerate}
\item Voice overlapping;
\item Low volume of the main speaker's voice, making the audio incomprehensible;
\item Word truncation.
\end{enumerate}

There are also 3 causes for considering a transcript as invalid, i.e. when it is not aligned with the audio, because there are:
\begin{enumerate}
\item Too many words in the transcript;
\item Too few words;
\item Words swapped.
\end{enumerate}



The following options were given to validate an audio/transcript pair:
\begin{enumerate}
\item Valid without problems.
\item Valid with filled pause(s).
\item Valid with hesitation.
\item Valid with background noise/low voice but understandable.
\item Valid with little voice overlapping.
\end{enumerate}

In cases where there is an audio with hesitation but the transcription does not correspond to the pauses made, the pair must be invalidated.
After one pair has been annotated, another is provided and this process continues until the user wants to stop the annotation and/or disconnect.




In the web interface for validation, the \textbf{transcription task} has a screen composed of the original transcription, a player for the audio file that can be repeated as many times as necessary, an editing window initially filled with the original transcription, which is used by the annotator to transcribe, and a button to send the transcription. To complete the task of transcribing an audio, the annotator must listen to the audio.

The annotator must also analyze if this audio fits into any of the types below: music, clapping, word truncation in the audio, loud noise or another language other than Portuguese, very low voice, incomprehensible voice, foul words, hate speech, and loud second voice. If so, the annotator should insert the symbols ``\#\#\#'' (denoting invalid audio) in the edit window and send its response. As we focused on the BP, we decided to kept 4.69 hours of European Portuguese, so during most of the project, annotators were instructed to discard European Portuguese audios.




The annotators were instructed to comply with the following eight guidelines:
\begin{enumerate}
\item Do not change to the grammar normative form the following signs of orality in the audio: ``tá/tó, né, cê, cês, pro, pra, dum, duma, num, numa''.
\item Transcribe filled pauses, such as ``hum, aham, uh'' as heard. 
\item Transcribe repetitive hesitations such as ``da da'', or ``do do'' as heard.
\item Write numbers in full form.
\item Letters that appear alone should be spelled out.
\item Acronyms and abbreviations should be transcribed in full form, using the English alphabet for those in English and the Portuguese alphabet for those that appear in Portuguese.
\item Foreign words should be transcribed normally, in the language in which they appeared.
\item Punctuation and case sensitivity could be applied, as normalization is performed in post-processing phase.
\end{enumerate}

\subsection{Kappa Evaluation: subjectivity of the Human Annotation} 
\label{sec:kappa}



The validation of audio-transcription pairs of the CORAA v1 corpus, using the binary annotation and transcription tasks (see Section \ref{sec:braz}), was performed from October 2020 to July 2021, when it was generated the database export.

The number of annotators varied during the project duration. In total, 63 different annotators performed the validation, which could be divided into 4 main annotation groups according to the start and end dates of each annotator on the project.
Two groups validated the corpora for 3 months in 2020 (October to December), with some annotators in this group continuing the validation in 2021. There was a 1-month annotation task-force during December 2020. The final group started the CORAA v1 validation work in May 2021 and ended in July 2021.



Each group attended a lecture on the validation process, read the tutorials for the two tasks (annotation and transcription) and received instructions to ask elucidate doubts via the project email throughout the process. 

At the beginning of the validation process, from October to December 2020, each audio-transcription pair was annotated by two or three annotators, so that we could use the majority vote to export the data, discarding the divergent pairs, in this initial phase of learning how to validate. Agreement between annotators was calculated in two ways: between annotators who annotated the same pairs (Section \ref{kappainicial}) and based on a gold-standard annotation of samples from all datasets, performed by a project member (Section \ref{kappafinal}).

\subsubsection{Kappa among Annotators}
\label{kappainicial}


Two Fleiss kappa values were calculated for the annotation from October to December 2020, to separate the groups of annotators. The project started with two groups in October, totaling 28 annotators, but with the entry of a new group on November 23, 2020 the number of annotators went to 63. Thus, it was decided to calculate a kappa value to evaluate each period of annotation --- from October 1st to November 23rd and from November 24th to December 31st, 2020. The hypothesis was that the annotation would become easier and with high agreement as the practice increased. However, there is another variable that influenced the agreement: the different transcription rules for each corpus of the CORAA corpus (see Section \ref{sec:text-norm}) also influenced the agreement. We calculated the agreement value via Fleiss' kappa twice, once considering only two annotators and the other considering only three annotators, according to the total number of annotators of a given audio. The values are shown in Table \ref{tab:kappavalues}. 

\begin{table*}[!ht]
\centering
\caption{Kappa values for each dataset in two annotation periods, separated by number of annotators. In the last three months of 2020, the order of validation of the corpora was C-Oral-Brasil I, SP-2010 and NURC-Recife.}

\begin{tabular}{l|l|l|l|l}
\hline
 & \multicolumn{2}{c|}{1/10 - 23/11}    & \multicolumn{2}{c}{24/11 - 31/12}    \\ \hline
                      & 2 annotators & 3 annotators  & 2 annotators  & 3 annotators \\ \hline
Number of pairs       & 6,785        & 29,835        & 26,974        & 4,224        \\ \hline
Number of annotators  & 25            & 25           & 51            & 51                   \\ \hline
\multicolumn{5}{c}{Kappa Values}                                                                                           \\ \hline
C-ORAL Brasil  I  & 0,394  & 0,353   & ---         & ---  \\ \hline
SP-2010          & 0,420  & 0,394   & ---         & ---   \\ \hline
NURC-Recife      & ---  & ---    & 0,317       & 0,314  \\ \hline
Total            & 0,391  & 0,392  & 0,317        & 0,314  \\ \hline
\end{tabular}
\label{tab:kappavalues}
\end{table*}

It is observed that there are absent values on the table, because the specific corpus was not being annotated in the referred period. The great disagreement between the annotators showed a more subjective task than previously imagined. By manually comparing audios in which annotators agreed with audios in which they disagreed, some points became clear: (i) the human ear naturally tends to complete truncated words, so that different annotators may disagree in defining whether an audio is in fact truncated or not, (ii) background noise level and voice pitch (low/high) are very subjective concepts, and different people are expected to consider different noise levels as tolerable, (iii) naturally, due to the ease of understanding different accents, annotators from different regions of the country tend to understand more or less of the audio according to the their accent, which can also be a source of disagreement.

\subsubsection{Kappa for the gold-standard annotation}
\label{kappafinal}




The gold standard was built to maintain the representativeness of all validated corpora, and all participating annotators, according to the following process:
\begin{enumerate}
\item For each annotated corpus, we generated a list of all annotators in that corpus;
\item For each name present in the list, five pairs annotated by the annotator were randomly selected (annotators with less than 5 pairs annotated per corpus had their pairs discarded);
\item The selected pairs were duplicated and annotated by an experienced annotator of the project, creating a gold-standard annotation with the following distribution:
\begin{itemize}
\item \textbf{Alip}: 15 annotators and 75 pairs
\item \textbf{C-ORAL Brasil I}: 24 annotators and 120 pairs,
\item \textbf{NURC-Recife}: 55 annotators and 275 pairs,
\item \textbf{SP-2010}: 25 annotators and 125 pairs,
\item \textbf{TEDx Portuguese}: 50 annotators and 250 pairs.
\item \textbf{Total}: 845 pairs (520 from the binary annotation task and 325 from the transcription task)
\end{itemize}
\end{enumerate}

The consensus pairs between the annotators were included in the exported dataset, that is, if the absolute majority chose to validate the pairs. Thus, we analyzed the degree of agreement of the annotators together (exported values) in comparison with the gold-standard corpus. The value obtained was \textbf{0.514}, showing a ``moderate agreement'', according to \cite{landis_kock_1977}. Even though the task is subjective, the final result obtained from the annotation of the exported pairs was satisfactory.
 
 
\subsection{Datasets Statistics} 
\label{sec:stat}

Overall, CORAA has 290.77 hours of validated audios, containing at least 65\% of its contents in the form of spontaneous speech. We will refer as the processed version of the corpora in CORAA as sub-datasets. NURC-Recife sub-dataset includes conference and class talks, considered prepared speech (see Table \ref{tab:original}).
Currently, no other dataset for BP includes audios in this speaking style. Therefore, the task of ASR is more challenging than for other datasets. Another CORAA characteristic is the presence of noise in some of its sub-datasets, which is also  more challenging for models created for this task. Table \ref{tab:datasets} presents statistics for each validated sub-dataset in CORAA v1. The resulting set encompasses almost 1,700 speakers. 

\begin{figure}[!h]
    \centering
    \includegraphics[width=1.0\columnwidth]{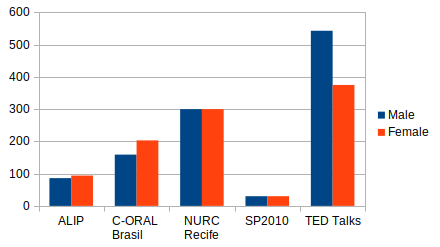}
    \caption{Estimated Speaker Distribution by Sex}
    \label{fig:sex_distribution}
\end{figure}
\begin{figure}[!h]
    \centering
    \includegraphics[width=1.0\columnwidth]{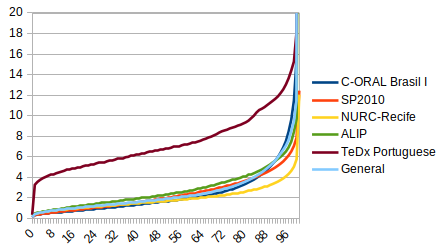}
    \caption{Duration distribution per sub-dataset}
    \label{fig:duration_percentil}
\end{figure}
\begin{figure}[!h]
    \centering
    \includegraphics[width=1.0\columnwidth]{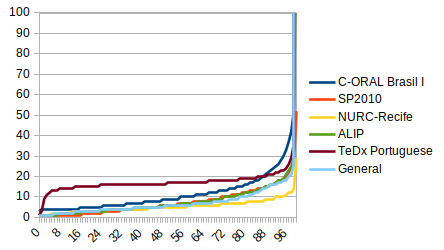}
    \caption{Word distribution per dataset}
    \label{fig:word_percentil}
\end{figure}

Audio durations range, in average, for 2.4 to 7.6 seconds according to sub-dataset. Audios having more than 200 words or 40 seconds were automatically filtered from the dataset. Figure \ref{fig:sex_distribution} presents estimated speaker distribution in  each sub-dataset according to sex. Overall, the distribution is similar for males and females\footnote{In the corpus C-ORAL Brasil I, there is a balance regarding number of uttered words --- 50.36\% words are uttered by 203 females and 49.64\% words are uttered by 159 males}. Figure \ref{fig:duration_percentil} presents audio duration distributions by sub-dataset. The audios are ranked by duration and their relative position (percentil) is shown in the $x$ axis. Audios duration are presented in the $y$ axis. Percentils are used to simplify sub-dataset comparisons. Figure \ref{fig:word_percentil} is similar, but presenting word distribution per dataset.

\begin{table*}[!ht]

\centering
\caption{Statistics for each processed version of the projects included in CORAA v1 (hours in decimal)}

\begin{tabular}{l|r|r|r|r|r|r}
\hline
\textbf{Corpus} &
  \textbf{ALIP} & \textbf{\begin{tabular}[c]{@{}l@{}}C-Oral \\ Brasil I\end{tabular}} &
  \textbf{\begin{tabular}[c]{@{}l@{}}NURC\\ Recife\end{tabular}} & \textbf{SP2010} &
  \textbf{\begin{tabular}[c]{@{}l@{}}TEDx\\ Port.\end{tabular}} &
  \textbf{Total} \\ \hline
Original (hrs)       & 78     & 21.13   & 279     & 65    & 249   & 692.21   \\ \hline
Validated (hrs)      & 35.96  & 9.64   & 141.31  & 31.14   & 72.74  & 290.79 \\ \hline
BP Speakers             & 179    & 362    & 417     & 60     & 671    & 1,689    \\ \hline
\begin{tabular}[c]{@{}l@{}}Audios \\ (segmented)\end{tabular} & 45,006 & 13,668 & 261,906 & 46,482 & 35,404 & 402,466 \\ \hline
\begin{tabular}[c]{@{}l@{}}Audio \\ Duration (sec.)\end{tabular}  & 2.90  & 2.46  & 1.94  & 2.43  & 7.55  & 3.39      \\ \hline
Avg Tokens     & 53.910    & 60.079    & 20.418     & 48.118   & 166.369    & 41.546      \\ \hline
Avg Types      & 6.391     & 7.188     & 3.733      & 6.002    & 8.807      & 5.581       \\ \hline
Total Tokens   & 335,664   & 99,954    & 1,378,558  & 339,890  & 610,639    & 2,764,705   \\ \hline
Total Types    & 14,189    & 8,715     & 41,903     & 12,351   & 27,469     & 58,237      \\ \hline
\begin{tabular}[c]{@{}l@{}}Type/Token  \\ Ratio \end{tabular}  & 0.042      & 0.087    & 0.030     & 0.036    & 0.046    & 0.022  \\ \hline
\end{tabular}
\label{tab:datasets}
\end{table*}

Regarding duration, the segmentation process  play a role in the obtained durations. Only ALIP and TEDx Portuguese were automatically segmented. The other sub-datasets were manually segmented. For the automatic segmentation, the parameters were adjusted aiming at better segmentation of informational units. ALIP had a similar duration than the others dataset. However, TEDx Portuguese audios tended to be longer. Speech style and genre also play a role in the obtained results. When pronounce is faster and with less pauses, there are less places in the audio that the segmentation software is confident to break the utterances. TEDx Portuguese is the main source of prepared speech in CORAA and had the longest audios and the same applies to word distribution, which is natural since the audios are longer. The remaining corpus presented similar distributions among them. 




\section{Baseline Model Development}
\label{sec:baseline}








{We performed a experiment over CORAA Dataset in order to measure the dataset quality, potentials and limitations. Before the execution of this experiment, the dataset was divided into three subsets: train, development and test. Table \ref{tab:datasets_training} presents the approximate number of hours for these sets for each sub-dataset, as well as the number of speakers from each sex. Sub-dataset validation sets were adjusted to have approximately 1 hour. Test sets were built in a similar manner, but having approximately 2 hours. This decision is supported by the work of \cite{sheshadri-etal-2021-wer}, which recommends that test sets should have at least 2 hours. NURC-Recife test set contains more than 3 hours of audios, because this sub-dataset have more speech genres than the others. All the audios from European Portuguese were included in the training set.}

\begin{table*}[!ht]
\centering
\caption{Statistics of Train/Dev/Test partitions of each CORAA corpora.}
\begin{tabular}{l|l|l|l|l|l|l}
\hline
\multicolumn{1}{l|}{\textbf{}} & \multicolumn{3}{l|}{\textbf{Duration (hrs)}} & \multicolumn{3}{l}{\textbf{Num. Speakers (M|F)}} \\ \hline
\multicolumn{1}{l|}{Subset}    & Train          & Dev          & Test         & Train            & Dev           & Test           \\ \hline
\multicolumn{1}{l|}{ALIP}      & 33.40         & 0.99       & 1.57        & 80|87              &  2|2            &  4|5               \\ \hline
\multicolumn{1}{l|}{C-ORAL Brasil I}     & 6.54           & 1.13         & 1.97 &  138|181 &      9|9     &  12|13               \\ \hline
NURC-Recife                         & 137.08        & 1.29         & 2.94         &   295|296               &      2|1         &      3|3 \\ \hline
SP2010                          & 27.83          & 1.13         & 2.18         &      27|27            &       1|1        &         2|2       \\ \hline
TEDx Portuguese                           &  68.67          & 1.37         & 2.70         &  532|364                &       4|4        &    7|7            \\ \hline
Total                           & 273.51         & 5.91         & 11.35        &        1072|955          &        18|17        &   28|30             \\ 
\hline
\end{tabular}
\label{tab:datasets_training}
\end{table*}


\subsection{Experiments}

{Our proposed experiment is based on the work of \cite{gris2021brazilian}. These authors fine-tuned the model Wav2Vec 2.0 XLSR-53 \citep{conneau2020unsupervised, baevski2020wav2vec} for ASR, using public available resources for BP. One of their experiments consisted on training 437.2 hours of Brazilian Portuguese. Wav2Vec 2.0 is model that learns quantized latent space representation from audios by solving a contrastive task. First, the model is pre-trained using an unsupervised approach in large datasets. Then, it is fine-tuned for the ASR task using supervised learning. Wav2Vec XLSR-53 is pre-trained over 53 languages, including Portuguese.}

In our approach, Wav2Vec XLSR-53 is fine-tuned for CORAA v1. We also evaluated \cite{gris2021brazilian} public fine-tuned model against CORAA v1, using the sets presented in Table \ref{tab:datasets_training}. 


Using the proposed training, development and testing divisions for CORAA v1, {we explores training Wav2Vec 2.0 XLSR-53 model using CORAA v1 during 40 epochs. Similarly to the work of \cite{conneau2020unsupervised} and \cite{gris2021brazilian}, we opted to freeze the model feature extractor.}


{To train the model, we use the framework HuggingFace Transformers \citep{wolf-etal-2020-transformers}. The model was trained with GPU NVIDIA TESLA V100 32GB using a batch size of 8 and gradient accumulation over 24 steps. We used the optimizer AdamW \citep{loshchilov2017decoupled} with a linear learning rate warm-up from 0 to 3e-05 in the first two epochs and after using linear decay to zero. During training the best checkpoint was chosen, using the loss in the development set. The code used to perform the experiment as well as the checkpoint of the trained model are publicly available at: \url{https://github.com/Edresson/Wav2Vec-Wrapper}.}

\subsection{Results and Discussions}
Section \ref{sec:comp:CV+Coraa} presents a comparison of our results with the work of \cite{gris2021brazilian}. The models are tested against the entire test subset of CORAA v1 and {Common Voice version 7.0 (Portuguese audios). Therefore, our model is evaluated \textit{in-domain} using CORAA v1 test set, a dataset in which it was fine-tuned for specific recording characteristics. At the same time, our model is also evaluated \textit{out-of-domain} in Common Voice, a dataset completely new to our model.}

Additionally, Section \ref{sec:comp:accents} focuses on evaluating the models in test sets of CORAA sub-datasets. This enables a more detailed analysis on factors such as audio quality and accents. Finally, Section \ref{sec:comp:spo-prep} investigates the two speech styles: prepared or spontaneous.  



\subsubsection{In/Out of Domain Evaluation} \label{sec:comp:CV+Coraa}

{Table \ref{tab:table-in-ou} presents the comparison of our experiment with the work of \cite{gris2021brazilian}. First, we performed an in-domain analysis of our model using CORAA v1 test set. Then, our model is evaluated out-of-domain using Common Voice test set. It is important to observe that, for the compared work, the analysis is mirrored, there is, CORAA v1 is the out-of-domain evaluation and Common Voice is the in-domain analysis.}

\begin{table*}[!ht]
\centering
\caption{Results for the In/Out of Domain Analysis.}
\begin{tabular}{c|c|c|c|c|c|c}
\hline
\textbf{Datasets}                        & \multicolumn{2}{c|}{\textbf{Common voice}}  &  \multicolumn{2}{c|}{\textbf{CORAA}}  & \multicolumn{2}{c}{\textbf{Mean}} \\ \hline
\textbf{Metric}                      & \textbf{CER} & \textbf{WER} & \textbf{CER}& \textbf{WER}  & \textbf{CER}& \textbf{WER}    \\ \hline
\cite{gris2021brazilian} & \textbf{4.15} & \textbf{13.85}& 22.32& 43.7&  13.23&  28.77  \\ \hline
Our     &6.34 & 20.08 & \textbf{11.02}& \textbf{ 24.18}& \textbf{8.68} & \textbf{22.13}    \\ \hline
\end{tabular}
\label{tab:table-in-ou}
\end{table*}

{In the Common voice dataset, as expected,  \cite{gris2021brazilian} model performed better. Regarding WER, it can be noted that our model is less than 7\% above their work. We also focuses our analysis on the metric CER, because for smaller audios, with just a few words, this metric tends to be more reliable. In this scenario, our models are approximately 2\% worse than the model from \cite{gris2021brazilian}. On the other hand, in the CORAA dataset, our model presented a much superior performance (more than 19\% in WER and 11\% in CER). Furthermore, our experiment managed to generalize better for audio characteristics not seen during training, achieving an average higher than the performance of the \cite{gris2021brazilian}. This is very interesting especially because the \cite{gris2021brazilian} model was trained with approximately 147 hours of speech more than our model.

We believe that models trained with the CORAA v1 dataset generalize better than a model trained with existing publicly available datasets for BP due to the spontaneous speech phenomenon and the wide range of noise and different acoustic characteristics present in CORAA. Furthermore, accent can be a factor since the datasets used in the training of the \cite{gris2021brazilian} model may not cover in depth all accents present in the CORAA v1. }


\subsubsection{Sub-dataset Analysis}\label{sec:comp:accents}

There are important differences in the recording environment for each sub-dataset. Additionally, they also varies on accents. Table \ref{tab:accent} presents the test performance for each CORAA v1 sub-dataset.

\begin{table*}[!ht]
\centering
\caption{Results in the CORAA test set for all subsets.}
\begin{tabular}{l|l|l|l|l}
\hline
\multirow{2}{*}{\textbf{Datasets}} & \multicolumn{2}{l|}{\textbf{\cite{gris2021brazilian}}} & \multicolumn{2}{l}{\textbf{Our}} \\ \cline{2-5} 
                                   & \textbf{CER}               & \textbf{WER}               & \textbf{CER}      & \textbf{WER}     \\ \hline
ALIP            &33.72 & 59.30 &17.30 & 34.06 \\ \hline
C-ORAL Brasil I & 23.53 & 45.9 & 13.62 &  28.88\\ \hline
NURC-Recife     &19.46 & 42.17 & 9.09 &  22.03 \\ \hline
TEDx Portuguese    &9.75 & 22.69& 7.43 & 19.36 \\ \hline
SP2010          & 23.11 & 42.44 & 9.57 & 20.00 \\ \hline
\end{tabular}
\label{tab:accent}
\end{table*}


Regarding datasets, ALIP presented the greatest challenge for the models, both for CER and WER metrics. We believe this occurred because audios from ALIP presented more noise than the other sub-datasets. 

Regarding accents, we have different results. On one hand, our model presented similar performances in NURC-Recife and SP2010, which have two distinct accents (Recife and São Paulo city). On the other hand, C-ORAL Brasil presented higher WERs and CERs than the other two. Two factors may have influenced this result. First, audio quality and noise presence tend to play a major role in model performances. Second, C-ORAL Brasil accent (Minas Gerais) has two characteristics that are difficult for models: speech rate is faster and there is more word agglutinations. As a consequence, the analysis was inconclusive for this accent, since the results are influenced both by the accent and the speech rate.

{
Regarding experiments, our model presented results varying from 19 to 34\% in WER and from 7 to 17\% in CER. On the other hand, \cite{gris2021brazilian} presented higher error rates, which is expected considering the training of their model had no previous contact with CORAA v1 audios. }

\subsubsection{Spontaneous vs Prepared Speech Analysis}\label{sec:comp:spo-prep}

Table \ref{tab:prepXspo} presents an analysis in which sub-datasets are merged according to speech style. The Spontaneous Speech column is obtained from the merging of ALIP, C-ORAL Brasil I, SP2010 and parts of NURC-Recife. The prepared speech column contains TEDx Portuguese and parts of NURC-Recife. As expected, the models perform better on prepared speech. However, for several ASR applications, spontaneous speech is more relevant (for example, ASR of phone call and meetings). This can also observed in Section \ref{sec:comp:accents}, as TEDx Portuguese presented the lowest error rates.

\begin{table*}[!ht]
\centering
\caption{Results Spontaneous vs Prepared Speech.}
\begin{tabular}{c|c|c|c|c}
\hline
\textbf{Speech Style}                      & \multicolumn{2}{c|}{\textbf{Spontaneous Speech}} & \multicolumn{2}{c}{\textbf{Prepared Speech}} \\ \hline
\textbf{Metric}                           & \textbf{CER}            & \textbf{WER}           & \textbf{CER}               & \textbf{WER}                \\ \hline
\cite{gris2021brazilian} & 25.75 & 49.18 & \textbf{5.30} &  \textbf{15.89 }                \\ \hline
 Our                     &  \textbf{12.44} & \textbf{26.5} & 6.07 &    18.7                  \\ \hline
\end{tabular}
\label{tab:prepXspo}
\end{table*}

\subsection{Error Analysis}
\label{sec:error}



The current test dataset is composed of 13,932 audio-transcription pairs, totaling 11.63 hours (see Section \ref{sec:baseline}), with parts from all CORAA v1 dataset.

{As this is the first time that a dataset composed of spontaneous speech samples was used to train an ASR model for BP, we performed a more detailed analysis of the errors from our model in a sample of the test dataset.}

The 13,932 test pairs were ordered by the CER values of our model to illustrate the different types of errors and to analyze whether there is a relationship of error types with CER values. The automatic transcription was analyzed using the typology of \cite{Mota_2000}, adapted for the task of evaluating ASR models.

The typology used here to illustrate the model errors is composed of 11 error types, grouped into 6 more general classes: Alphabetical, Lexical, Morphological, Language and Spontaneous Speech, Semantic, and Diacritic Placement Errors. Below we present a description of the 11 errors with examples.

\begin{itemize}
\item Alphabetical errors are alphabetic writing application errors. \\
1) Alphabetical errors occur in 3 situations: by transcribing speech directly into writing, in complex syllables or even
with ambiguous letters (``ce'' versus ``sse'' or ``sa'' versus ``za'', in Portuguese). \\
An example of this type of error is related to the sound /k/ in Portuguese which is represented by the letter ``c'' before some vowels and by ``qu'' before other vowels. Thus, the use of ``c'' in place of ``qu'' is associated with the speaking/writing relationship.
\item Lexical errors occur in an excerpt transcribed by the ASR where there is:\\
2) omission or addition of words; \\ 3) exchange of words. \\
An example from our dataset regarding addition of a word in the automatic transcription is 	``que legal'' instead of ``legal'' \\ 
Also from our dataset, an example of word exchange is ``e que mais que a gente vida'' instead of ``e que mais que a gente viu''.
\item Morphological errors are errors that occur due to the violation of writing rules that is linked to the morphological structure of words. These are errors from: \\
4) omitting morphemes (e.g. ``come'' written instead of ``comer'');\\
5) concatenation of morphemes (e.g., ``agente'' instead of ``a gente'', or ``acasa'' instead of ``a casa'' );\\
6) separation of morphemes, as in the example: ``de ele''  written instead of the contraction ``dele'').
\item Language and spontaneous speech errors are errors of: \\
7) Words in English (or in a language other than BP) wrongly transcribed;\\
8) Filled pause errors (e.g., ``é'' versus ``eh'' ) where the transcription and model responses diverge; \\
9) Spontaneous speech errors (e.g., ``tá'' versus ``está'';  ``té'' versus ``até''; ``cê'' versus ``você'') in which transcription and model responses diverge.
\item Semantic errors occur when two words are spelled similarly but have different meanings.\\
10) Semantic errors (e.g. ``Ela comprimentou o diretor assim que chegou.'', where the correct form would be ``cumprimentou'').
\item Diacritic placement errors occur due to missing accents or improperly adding them. They are problematic because the five training corpora were built at different times, in which there were different spelling rules for the Portuguese language. For example, the last orthographic agreement for the Portuguese language came into force in Brazil in 2016.\\
11) Accent marks errors.
\end{itemize}

Table \ref{tab:exemplos_erros} shows examples of 11 errors presented above (column 1), in which the original transcript (column 2) and the model response (column 3) diverge. The location of the error in the snippets appears in bold.

\begin{table*}[ht]
\centering
\caption{Examples of the 11 different error types.}
{\small
\begin{tabular}{l|l|l}
\hline
 \begin{tabular}[c]{@{}c@{}}Error \\ Type\end{tabular} & Original Transcription & ASR Transcription  \\
\hline
1          & \begin{tabular}[c]{@{}l@{}}uma maneira de saber o que e \\ como o indivíduo \textbf{identifica} algo\end{tabular}          & \begin{tabular}[c]{@{}l@{}}uma maneira de saber o que e \\ como o indivíduo \textbf{identifiqa} algo\end{tabular}            \\
\hline
2          & ou pra dar um apoio moral   & ou pra dar um apoio \textbf{im} moral        \\                            
\hline
3          & \begin{tabular}[c]{@{}l@{}}o outro \textbf{foi} morar um pouco \\ mais longe\end{tabular}                                                 & \begin{tabular}[c]{@{}l@{}}o outro \textbf{prai} morar um pouco \\ mais longe\end{tabular}                                                                              \\
\hline

5,4        & \textbf{que lhe dão} ora \textbf{dor}                                                                                              & \textbf{ciridão} ora \textbf{do}                                                                                                     \\
\hline
5          & criança é mais \textbf{coca cola}* biscoito                                                                                 & criança é mais \textbf{cocacola} biscoito                                                                                    \\
\hline
6          & \begin{tabular}[c]{@{}l@{}}que levaria a uma resposta \\ \textbf{aquele} estímulo\\ \\ ah legal faz \textbf{tempão} já\end{tabular} & \begin{tabular}[c]{@{}l@{}}que levaria a uma resposta \\ \textbf{a quele} estímulo\\ \\ ah legal faz \textbf{tem pão} já\end{tabular} \\
\hline
7          & \begin{tabular}[c]{@{}l@{}}na teoria de \textbf{osgood} é que\\ \\ de \textbf{jazz}\end{tabular}                              & \begin{tabular}[c]{@{}l@{}}na teoria de \textbf{osguot} é que\\ \\ de \textbf{dez}\end{tabular}     \\             
\hline
8          & \begin{tabular}[c]{@{}l@{}}e essa daí \textbf{eh}\\ \\ \textbf{eh}\\ \\ \textbf{eh}\\ \\ \textbf{ham}\end{tabular}               & \begin{tabular}[c]{@{}l@{}}e essa daí \textbf{é}\\ \\ \textbf{é}\\ \\ \textbf{ahn}\\ \\ \textbf{uhn}\end{tabular}                                       \\
\hline
9          & \begin{tabular}[c]{@{}l@{}}\textbf{pra} área específica que é o curso \\ diz que é um curso excelente\end{tabular}         & \begin{tabular}[c]{@{}l@{}}\textbf{para} área específica que é o curso \\ diz que é um curso excelente\end{tabular}          \\
\hline
10         & \begin{tabular}[c]{@{}l@{}}entendeu era eles \textbf{suavam} mais a \\ camisa pelo clube entendeu e\end{tabular}           & \begin{tabular}[c]{@{}l@{}}entendeu era eles \textbf{soavam} mais a \\ camisa pelo clube entendeu e\end{tabular}             \\
\hline
11         & \begin{tabular}[c]{@{}l@{}}então \textbf{é} conhecer a população \\ usuária do equipamento urbano\end{tabular}             & \begin{tabular}[c]{@{}l@{}}então \textbf{e} conhecer a população \\ usuária do equipamento urbano\end{tabular} \\
\hline
\end{tabular}
}
\label{tab:exemplos_erros}
\\
** The lack of a hyphen in the test  set is only for the calculation of CER/WER.
\end{table*}

A sample of 938 audio-transcription pairs was analyzed, of which 134 contained some errors in the audio transcription and thus they were not framed in the typology. Also, 309 pairs were annotated for deletion as their audio were compromised (because of truncation, very loud noise or overlapping voices). The remaining 500 pairs, according to the CER ranges analyzed, are shown in column 1 of Table \ref{tab:CER_erros}. They were categorized according to the typology presented above. For some pairs more than one error occurs and for some excerpts with high CER values only one error was annotated (the most frequent) although the transcription had many more.

This initial analysis has already resulted in a decision to make a revision in all pairs of the test dataset, which is currently being conducted, and should result in a new version CORAA in the future.

Table \ref{tab:CER_erros} shows, in the last column, the variety of error types in each range presented column 1; its frequency is shown in parentheses. We present in bold the most frequent type.


\begin{table*}[ht]
\caption{Intervals of CER and frequencies of the different error types.}
\centering
\begin{tabular}{l|l|c|l}
\hline
Intervals         & CER       & \begin{tabular}[c]{@{}c@{}}Analysed \\ Samples\end{tabular} & Error Types (occurrences)                                                                                         \\ \hline
1      --- 4,613  & 0          & ---                                                             & ---                                                                                                          \\ \hline
4,614  --- 8,397  & $0 <$ CER $< 0.1$    & 110                                                            & \begin{tabular}[c]{@{}l@{}}1 (1), 2 (3), 4 (5), \textbf{5 (66)}, 6 (28), \\ 8 (1), 9 (1), 10 (1), 11 (2)\end{tabular} \\ \hline
8,398  --- 10,724 & $0.1 \leq$ CER $< 0.2$    & 10                                                             & 2 (3), \textbf{3 (8)}, 4 (1), 5 (1), 7 (1)                                                                            \\ \hline
10,725 --- 11,991 & $0.2 \leq$ CER $< 0.3$    & 10                                                             & 2 (2), \textbf{3 (7)}, 8 (1)                                                                                          \\ \hline
11,992 --- 12,666 & $0.3 \leq$ CER $< 0.4$    & 10                                                             & 2 (2), \textbf{3 (14)}, 4 (1), 5 (3), 9 (1)                                                                           \\ \hline
12,667 --- 13,049 & $0.4 \leq$ CER $< 0.5$    & 25                                                             & 2 (5), \textbf{3 (29)}, 4 (2), 5 (9), 6 (1), 8 (1)                                                                    \\ \hline
13,050 --- 13,336 & $0.5 \leq$ CER $< 0.6$    & 10                                                             & 2 (2), \textbf{3 (6)}, 6 (1), 7 (1), 9 (1)                                                                            \\ \hline
13,337 --- 13,509 & $0.6 \leq$ CER $< 0.7$    & 10                                                             & 2 (2), \textbf{3 (5)}, 5 (2), 6 (2), 8 (1)                                                                            \\ \hline
13,510 --- 13,932 & $0.7 \leq$ CER $\leq 12$ & 315                                                            & \textbf{3 (42)}, 6 (2), 8 (35), 9 (2)                                                                                 \\ \hline
\end{tabular}
\label{tab:CER_erros}
\end{table*}

The lexical error of type 3 --- exchange of words --- is the most frequent one, which is expected given that the task is automatic transcription, and the training process of these models favors the recognition of frequent and well formed words. 
Moreover, omission and addition of words (error type 2) is pervasive as it appears in all the intervals (even in the last one, where CER varies from 0.7 to 12, although it was not  explicitly annotated). 
However, the second and third errors classified by frequency are: concatenation and filled pause swap error. The latter is related to the fact that the CORAA dataset has a large percentage of spontaneous speech samples in which both the number and variety of filled pauses are high.

After this error analysis, it became clear the need for more normalization rules for filled pauses representations so that the model accuracy increases. 

\section{Conclusions and Future Work}
 \label{sec:conc}


In this paper we presented and made publicly available a new dataset called CORAA v1, with 290.77 hours of validated pairs of audio-transcription, composed by public corpora in BP and TEDx Talks in European and Brazilian Portuguese.

Counting on the cooperation among research centers, universities, private companies and The São Paulo Research Foundation (FAPESP), we made publicly available this new and large dataset for training BP speech recognition models, closing the gap of the previous datasets, i.e., the lack of spontaneous and informal speech used in conversations, dialogues and interviews. 
Informed by the error analysis, we are normalizing filled pauses representations and performing a new validation over the test and development datasets, in order to increase future model accuracy. 

As for future work, we plan to augment CORAA with new corpora from Tarsila Project\footnote{\url{https://sites.google.com/view/tarsila-c4ai}} such as Museu da Pessoa\footnote{\url{https://museudapessoa.org/}} and NURC-SP\footnote{\url{https://nurc.fflch.usp.br/}}. We also plan to create an ASR Challenge including CORAA v1 to further develop research in ASR for the Portuguese language, 
in order to motivate young researchers in this exciting research area. Finally, we plan to refine the normalization rules of filled pauses and deliver a new version of CORAA dataset.

\section{Acknowledgements}

  This research was funded by CEIA with support by the Goi\'{a}s State Foundation (FAPEG grant \#201910267000527)\footnote{\url{http://centrodeia.org/}}, Department of Higher Education of the Ministry of Education (SESU/MEC), Copel Holding S.A.\footnote{\url{https://www.copel.com}}, and Cyberlabs Group\footnote{\url{https://cyberlabs.ai/}}. In addition, this research was carried out at the Center for Artificial Intelligence (C4AI-USP), with support by the São Paulo Research Foundation (FAPESP grant \#2019/07665-4) and by the IBM Corporation.
  Also, this study was financed in part by the Coordena\c{c}\~{a}o de Aperfei\c{c}oamento de Pessoal de N\'{i}vel Superior -- Brasil (CAPES) -- Finance Code 001. We also would like to thank Nvidia Corporation for the donation of Titan V GPU used in CORAA related projects. The coauthor Anderson da Silva Soares thanks to CNPq for Productivity Scholarship in Technological Development and Innovative Extension - number 308808/2020-7. Finally, the authors would like to thank all the members of the TARSILA project that contributed with discussions and insights regarding the compilation of CORAA v1 corpus. 

\bibliographystyle{IEEEtran}

\bibliography{mybib}


\end{document}